\documentclass{article}

\usepackage[final]{neurips_2025}

\usepackage[utf8]{inputenc} % allow utf-8 input
\usepackage[T1]{fontenc}    % use 8-bit T1 fonts
\usepackage{hyperref}       % hyperlinks
\usepackage{url}            % simple URL typesetting
\usepackage{booktabs}       % professional-quality tables
\usepackage{amsfonts}       % blackboard math symbols
\usepackage{amsmath}
\usepackage{nicefrac}       % compact symbols for 1/2, etc.
\usepackage{microtype}      % microtypography
\usepackage{xcolor}         % colors
\usepackage{algorithm}
\usepackage{algorithmic}
\usepackage{subcaption}
\usepackage{graphicx}
\usepackage{listings}
\usepackage{cleveref}
\usepackage{tikz}
\usepackage{amssymb}
\usepackage{xcolor}
\usepackage{hyperref}
\usepackage{fontawesome}

\usepackage{makecell}

\definecolor{ggreen}{HTML}{68DB7C}
\definecolor{oorange}{HTML}{FFA94D}
\definecolor{bblue}{HTML}{4DABF7}

\newcommand{\primerl}{\texttt{prime-rl}}
\newcommand{\INTELLECT}{\textsc{INTELLECT-3}}
\newcommand{\verifiers}{\texttt{verifiers}}
\newcommand{\envhub}{Environments Hub}

\newcommand{\torchtitan}{\texttt{torchtitan}}

\newlength{\myMheight}
\newcommand{\github}{\includegraphics[height=\myMheight]{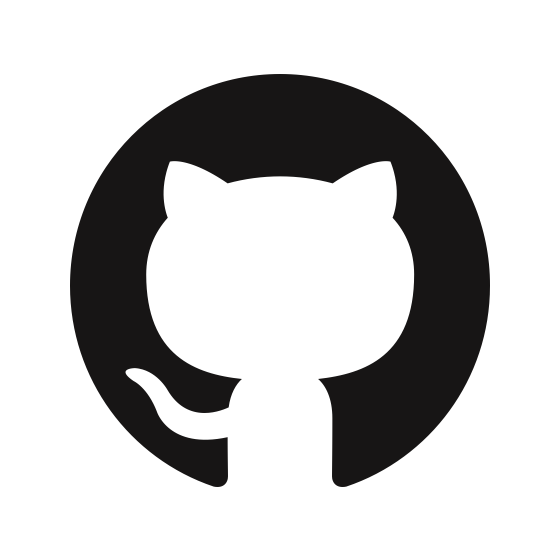}}
\newcommand{\hf}{\includegraphics[height=\myMheight]{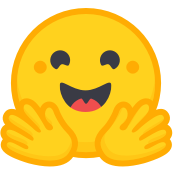}}
\newcommand{\pilogo}{\includegraphics[height=\myMheight]{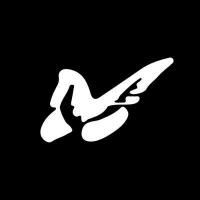}}

\title{\INTELLECT: Technical Report}

\author{%
  Prime Intellect Team \\[6pt]
  Mika Senghaas \quad Fares Obeid \quad Sami Jaghouar \quad William Brown \\
  Jack Min Ong \quad Daniel Auras\thanks{Partially while at \texttt{ellamind}} \quad
  Matej Sirovatka \quad Jannik Straube \\
  Andrew Baker \quad Sebastian Müller \quad Justus Mattern \quad Manveer Basra \\
  Aiman Ismail \quad Dominik Scherm \quad Cooper Miller \quad Ameen Patel \\
  Simon Kirsten \quad Mario Sieg \quad Christian Reetz \quad Kemal Erdem \\
  Vincent Weisser \quad Johannes Hagemann\thanks{Prime Intellect, Inc. Correspondence to: \texttt{johannes@primeintellect.ai}}%
}

\begin{document}

{
\begingroup
\begin{figure*}
    \centering
    \hspace{0.32cm}
    \includegraphics[width=0.125\textwidth]{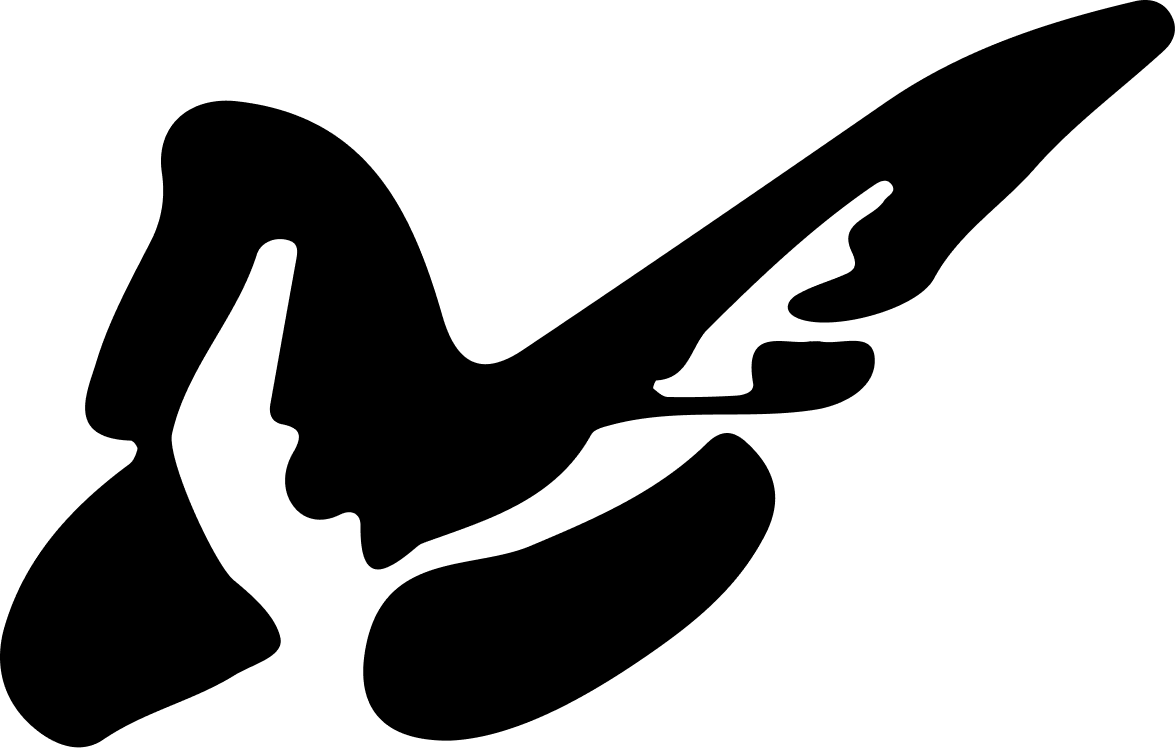}
\end{figure*}
\endgroup
}

\maketitle

\begin{abstract}
We present \INTELLECT, a 106B-parameter Mixture-of-Experts model (12B active) trained with large-scale reinforcement learning on our end-to-end RL infrastructure stack. \INTELLECT\ achieves state of the art performance for its size across math, code, science and reasoning benchmarks, outperforming many larger frontier models.
We open-source the model together with the full infrastructure stack used to create it, including RL frameworks, complete recipe, and a wide collection of environments, built with the \verifiers\ library, for training and evaluation from our \envhub\ community platform.

Built for this effort, we introduce \primerl, an open framework for large-scale asynchronous reinforcement learning, which scales seamlessly from a single node to thousands of GPUs, and is tailored for agentic RL with first-class support for multi-turn interactions and tool use. Using this stack, we run both SFT and RL training on top of the GLM-4.5-Air-Base model, scaling RL training up to 512 H200s with high training efficiency.
\end{abstract}

\setcounter{footnote}{0}
\begin{figure}[htb]
\centering
\includegraphics[width=0.8\textwidth]{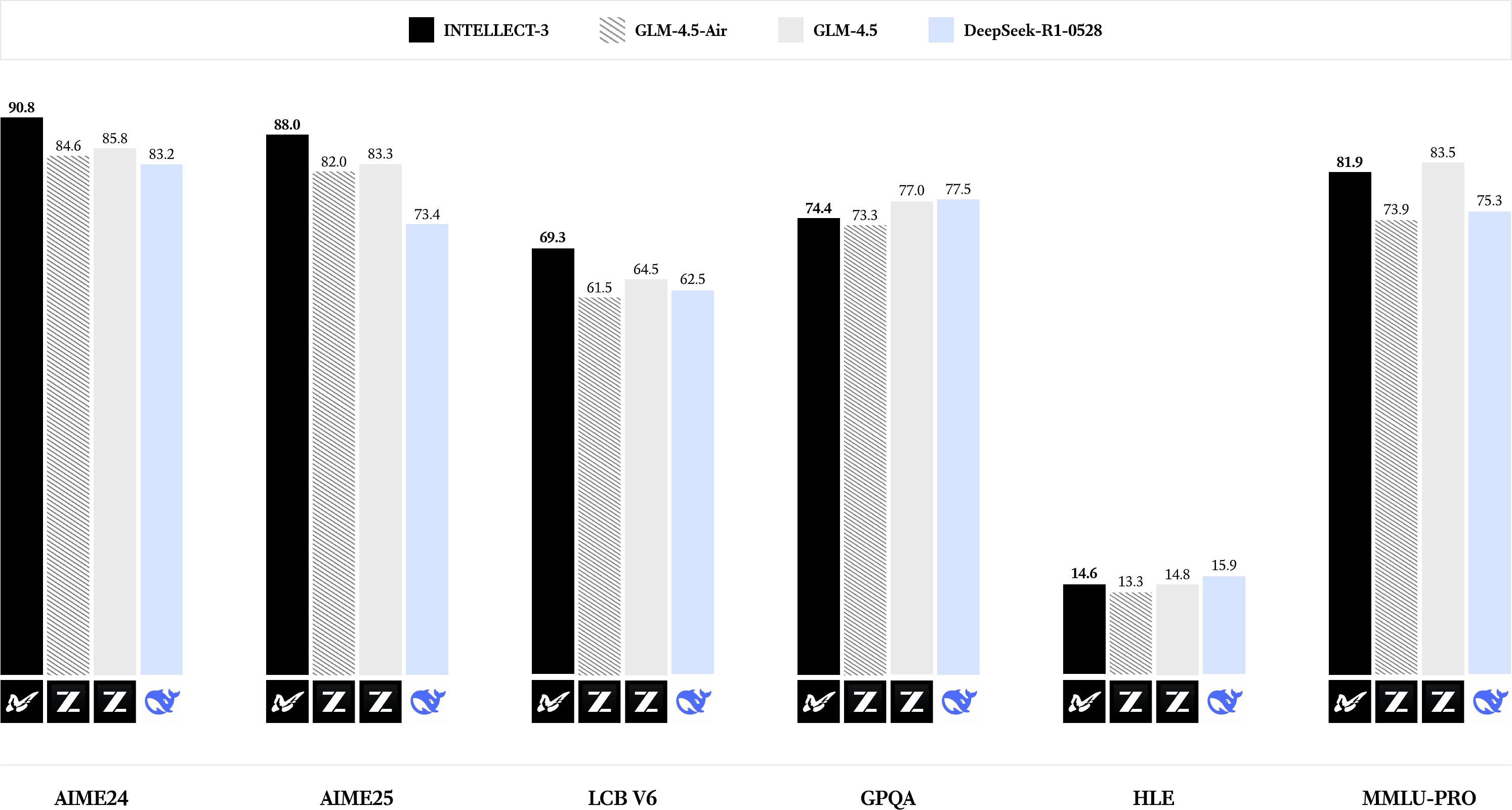}
\captionsetup{width=0.9\textwidth}
\caption{\centering \INTELLECT\ Evaluation Results.\protect\footnotemark}
\end{figure}

\newpage

\footnotetext{All the models above were evaluated using our public, reproducible \envhub\ implementations. To ensure the best results, we use APIs provided directly by the model creators whenever available to avoid any performance loss from quantization or other inference optimizations.}

\tableofcontents

\newpage

\section{Introduction}
\label{sec:introduction}

Scaling compute for training large language models (LLMs) with reinforcement learning with verifiable rewards (RLVR) has emerged as the dominant paradigm for improving model performance in post-training. Models such as OpenAI o3~\cite{openai2025o3}, Grok 4~\cite{xai2025grok4}, and DeepSeek R1~\cite{deepseekai2025deepseekr1} demonstrate that training models via RL for long-context reasoning and agentic tool use greatly enhances their capabilities, making them more effective both for everyday and specialized tasks.

While the open-source ecosystem has been successful in producing strong open-weight models trained with RL, the open-source infrastructure for the end-to-end RL pipeline, including training frameworks, RL environments and evaluations, and stable training recipes, still has notable shortcomings compared to proprietary pipelines inside frontier AI labs. For example, existing open-source frameworks are often complex, monolithic, and designed without modularity in mind~\cite{sheng2024hybridflow}. This can make extensibility difficult, inhibit broad adoption, slow down individual research projects, and lead to a fragmentation of ecosystem artifacts.

In this report we present \INTELLECT, a state-of-the-art model in its weight class built on top of the GLM-4.5-Air base model~\cite{5team2025glm45agenticreasoningcoding}. Alongside the model, we share the complete training recipe, covering everything from supervised fine-tuning on the initial base checkpoint to large-scale reinforcement learning. We also introduce our training framework \primerl, which is easy to use and hackable yet performant and scalable enough to support state-of-the-art RL post-training. We highlight the following features:

\begin{enumerate}
    \item First-class support for OpenAI-compatible async inference, \verifiers\ environments~\cite{brown2025verifiers}, and a public \envhub\ to standardize agentic RL training and evaluation
    \item Support for end-to-end post-training, including SFT and multi-turn agentic RL
    \item Multi-node deployment with FSDP2 training and vLLM inference backend
    \item Naturally asynchronous training for high-throughput including continuous batching and in-flight weight updates~\cite{piché2025pipelinerlfasteronpolicyreinforcement}
    \item Modular and extensible by nature, enabling high research velocity
\end{enumerate}

\INTELLECT\ fully trained end-to-end with \primerl\ and our open-source infrastructure components, outperforms existing open-source models in its size range across the board and even surpasses frontier open models over $6\times$ larger on reasoning and agentic benchmarks: It achieves scores of 90.8\% and 88.0\% on AIME 2024 and 2025 respectively, outperforming DeepSeek's frontier models and matching the performance of Z.ai's latest next-generation model GLM-4.6 which has over $3\times$ the number of parameters. On coding benchmarks, \INTELLECT\ achieves 69.3\% on LiveCodeBench v6, outperforming Z.ai's GLM-4.5-Air post-train by 8\%.

We open-source \INTELLECT\footnote{\hf ~\href{https://huggingface.co/PrimeIntellect/INTELLECT-3}{\texttt{huggingface.co/PrimeIntellect/INTELLECT-3}}}, our RL training framework \primerl\footnote{\github ~\href{https://github.com/PrimeIntellect-ai/prime-rl}{\texttt{github.com/PrimeIntellect-ai/prime-rl}}}, and all environments~\footnote{\pilogo ~\href{https://hub.primeintellect.ai}{\texttt{hub.primeintellect.ai}}} used for synthetic data generation, training, and evaluation.

The remainder of this report is organized as follows: \Cref{sec:infra} provides a detailed overview of the end-to-end reinforcement learning infrastructure, including \primerl, \verifiers, the \envhub, sandbox code execution and compute orchestration. \Cref{sec:intellect-3-training} describes the concrete \INTELLECT\ training run, covering the RL environments used for training and the results of both the SFT and RL stages. \Cref{sec:evaluations} presents the model’s evaluation results on reasoning and agentic benchmarks. Finally, \Cref{sec:summary} concludes the report and outlines directions for future work.

\section{Training Infrastructure}
\label{sec:infra}

We introduce the following key training infrastructure components for the end-to-end training of \INTELLECT\:

\begin{itemize}
    \item \textbf{prime-rl:} An asynchronous RL framework powering large-scale SFT and RL training of Mixture-of-Experts models.
    \item \textbf{Verifiers and the \envhub:} A unified environment interface and ecosystem for agentic RL.
    \item \textbf{Prime Sandboxes:} High-throughput, secure code execution for agentic coding environments.
\end{itemize}

\subsection{prime-rl: A Framework for Asynchronous Reinforcement Learning at Scale}
\label{sec:design-architecture}

\INTELLECT\ was trained end-to-end with \primerl, our production-scale post-training framework. \primerl\ provides native integration with \verifiers\ environments, which power our entire post-training stack from synthetic data generation, supervised fine-tuning, reinforcement learning, to evaluations. Through its tight connection to the \envhub, the entire training stack can seamlessly access a rapidly expanding ecosystem of training and eval environments.

\subsubsection{Architecture}
\label{subsec:architecture}

Three main abstractions facilitate RL training: the {\it orchestrator}, the {\it trainer}, and the {\it inference} service.

\begin{figure}[ht]
\centering
\includegraphics[width=\textwidth]{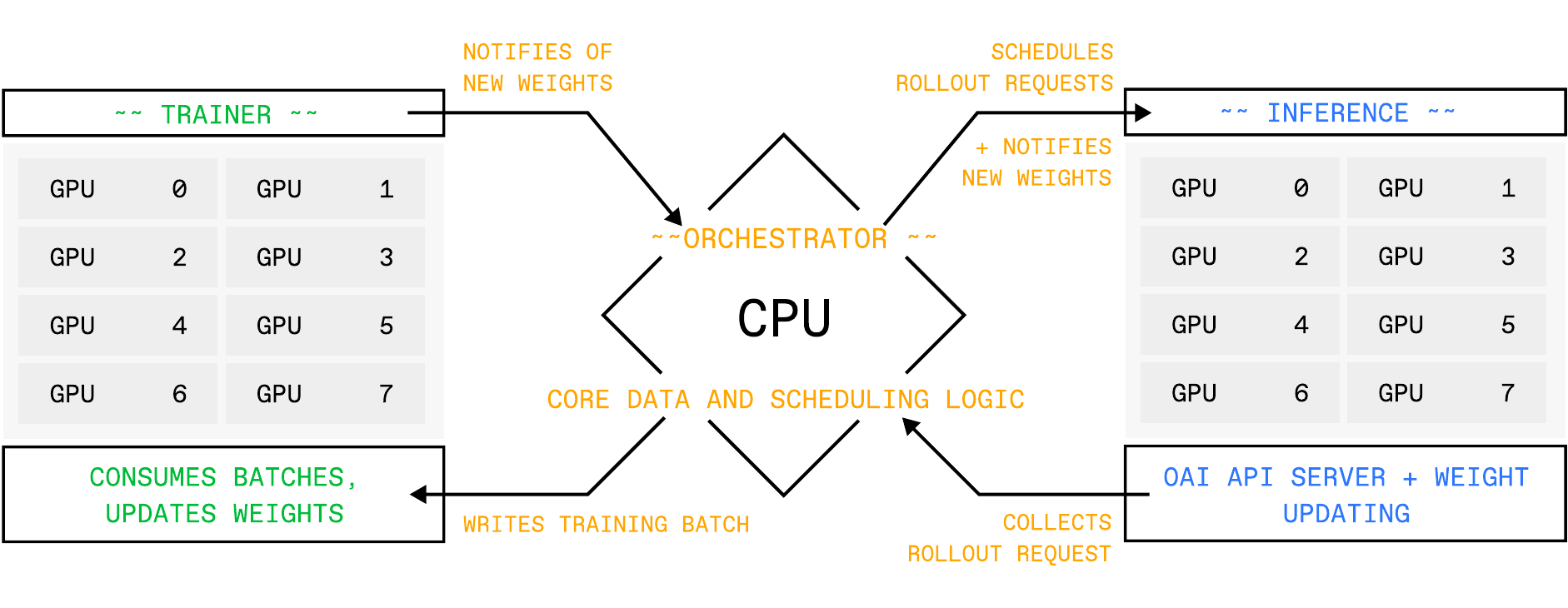}
\captionsetup{width=0.9\textwidth}
\caption{\centering \textbf{Architecture.} A RL training run involves the coordination of a \textcolor{ggreen}{trainer}, \textcolor{oorange}{orchestrator} and an \textcolor{bblue}{inference} service. The FSDP trainer and vLLM inference run disaggregated, and can be individually deployed across multiple nodes.}
\end{figure}

\textbf{Orchestrator.} The orchestrator is a lightweight CPU process that handles the core data flow and scheduling logic, serving as an intermediary between the trainer and inference service with bidirectional relays. In one direction, it collects rollouts from the inference server, assembles them into packed batches, and dispatches them to the trainer; in the other direction, it relays updated model weights from the trainer to the inference service. The orchestrator utilizes \verifiers\ environments to abstract multi-turn rollout generation and scoring, allowing any environment on the \envhub\ to plug into the training loop.

\textbf{Trainer.} The trainer is responsible for producing an updated policy model given rollouts and advantages. We use FSDP 2~\cite{pytorch2025} as the backend with compatibility for any HuggingFace (HF) model. FSDP shards model parameters, gradients, and optimizer states, allowing training large models with data parallelism and minimal GPU memory footprint. The trainer is inspired by \torchtitan~\cite{liang2025torchtitan} and relies on native PyTorch features to implement advanced parallelism techniques, such as tensor, context, and expert parallelism, and leverages grouped matrix multiplication kernels for efficient MoE training.

\textbf{Inference.} The inference pool consists of standard OpenAI-compatible servers with a vLLM~\cite{kwon2023vllm} backend. The API specification is extended with custom endpoints to enable updating the server with the latest policy: \texttt{/update\_weights} is used to update the policy, and \texttt{/reload\_weights} is used to reset the weights to the base model in between experiments. We rely on vLLM's optimized kernels, parallelism strategies, and scheduling for fast rollout generation. Given the disaggregated nature of the service architecture, it can be directly extended to include multiple engines with a shared request pool, allowing operation across multiple clusters and straightforward integration of alternative inference engines (e.g. SGLang~\cite{zheng2024sglang}, Tokasaurus~\cite{juravsky2025tokasaurus}).

\subsubsection{Asynchronous Off-Policy Training}
\label{subsec:async-off-policy-training}

In \primerl, the trainer and inference run disaggregated, i.e. on a disjoint set of GPUs, to overlap rollout generation and training.

At each training step, all artifacts are identified by the step count $n$. For the trainer, this is the gradients $g_n$ and model weights $\theta_n$, and for the inference service, rollouts $(x_n, y_n)$. At step 0, the inference service uses $\theta_0$ (base model) to produce $(x_0, y_0)$. The trainer subsequently uses $(x_0, y_0)$ to compute $g_0$ to finally update the model as $\theta_1 \leftarrow \theta_0 - g_0$.

\begin{figure}[ht]
\centering
\vspace{8pt}
\includegraphics[width=\textwidth]{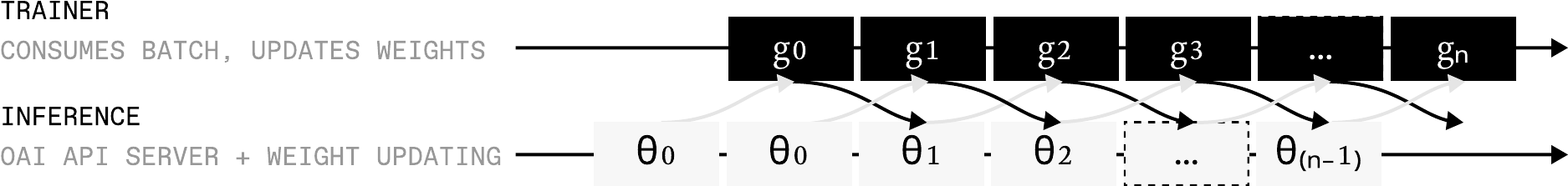}
\vspace{4pt}
\caption{\centering \textbf{Asynchronous Off-Policy Training.} We show the execution graph of one-step off-policy training in an idealized setting where the trainer step time equals the inference step time. At step $n$, the inference engine uses a policy no older than $\theta_{\min{(0,n-1)}}$.}
\label{fig:asynchronous-off-policy-training}
\end{figure}

In synchronous on-policy training, the inference engine stalls after producing $(x_0, y_0)$ because it requires $\theta_1$ to produce the next rollouts $(x_1, y_1)$. To prevent this, we allow off-policy training, which means that the inference service can asynchronously generate rollouts from an old policy model. For example, in one-step off-policy training, the inference service continues generating $(x_1, y_1)$ from $\theta_0$, while the trainer is producing $\theta_1$ in parallel. An example of such overlapped, asynchronous computation is shown in ~\autoref{fig:asynchronous-off-policy-training}. 

\subsubsection{Continuous Batching \& In-Flight Weight Updates}
\label{subsec:continuous-batching}

Generating rollouts which are many tens of thousands of tokens long quickly becomes a major bottleneck in large-scale RL training. Traditional systems schedule $n$ rollout requests and only release a training batch once the slowest rollout has finished generation. However, this method is prone to heavily under-utilize inference compute because less-than-optimal rollouts are being generated concurrently as rollouts in the batch finish. This bottleneck becomes especially visible if there is high variance in the length of the generated rollouts, as is typical for reasoning models in complex agentic environments.

To make training viable, we implement continuous batching with in-flight policy weight updates, as popularized by AReal~\cite{fu2025areallargescaleasynchronousreinforcement} and PipelineRL~\cite{piché2025pipelinerlfasteronpolicyreinforcement}.
Two main asynchronous task loops on the orchestrator achieve this.

\textbf{In-Flight Weight Updates.} The orchestrator continuously polls the trainer to update the inference pool as soon as a new policy becomes available. Once this happens, the inference pool temporarily interrupts generation to receive the updated weights from the trainer. Once the weight update is complete, rollout generation continues with the updated weights. Thus, a single trajectory may be generated by multiple policies. We control the maximum off-policyness by discarding rollouts which are generated by more than \texttt{max\_off\_policy\_steps} policies to prevent policy drift.

\textbf{Continuous Batching.} The orchestrator maintains a large pool of concurrent, asynchronous rollout requests. Whenever a rollout group completes, its slot is immediately repopulated with a new request. This keeps the pool saturated and ensures that a constant number of rollouts are in flight, thereby sustaining peak inference throughput without waiting for synchronous batch boundaries.

\begin{figure}[ht]
\centering
\includegraphics[width=0.85\textwidth]{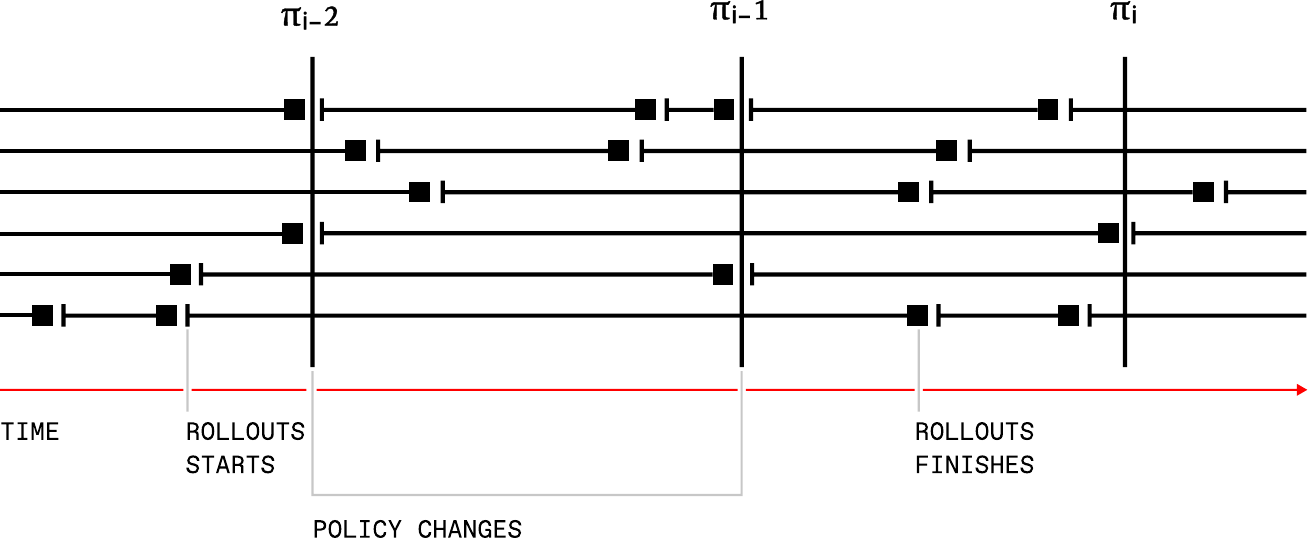}
\captionsetup{width=0.9\textwidth}
\caption{\centering \textbf{Continuous Batching \& In-Flight Weight Updates.} Continuous batching maintains a constant inference load because finished rollout are immediately replaced with new rollout requests. The policy used to generate rollouts is updated in-flight as soon as it becomes available, allowing rollouts to be generated by multiple policies.
}
\label{fig:pipelinerl}
\end{figure}

Figure~\ref{fig:pipelinerl} visualizes how rollout trajectories span multiple policies which are updated in-flight as they become available.

We find these optimizations critical to scale to long-context RL training, achieving much higher end-to-end system throughput, while minimizing data off-policyness.

\subsubsection{Multi-Client Orchestrator}
\label{subsec:multi-client-orchestrator}

When scaling inference to hundreds of GPUs, we found that the standard multi-node data-parallel strategy provided by vLLM did not deliver the expected throughput gains. As the number of inference nodes increased, overall performance quickly plateaued.

To overcome this limitation, we implemented a custom data-parallel strategy centered on a multi-client abstraction on the orchestrator. Each inference node is deployed as an entirely independent server, and the orchestrator maintains one client per node. Group rollout requests are distributed across clients using a simple round-robin mechanism, ensuring balanced utilization and eliminating any inter-node synchronization. We found that inference throughput scales linearly with the number of nodes used, as desired.

\subsubsection{Online Data Filtering}
\label{subsec:online-data-filtering}

An effective reinforcement learning setup depends on a well-designed curriculum that exposes the model to tasks of appropriate and progressively increasing difficulty. Beyond pre-selecting data via offline data filtering, we implement advanced online data filtering techniques to continually adapt task difficulty during training.

To support this, problems are sorted into difficulty pools (easy, normal, hard) based on each problem’s observed solve rate. By flexibly controlling how many samples are drawn from each pool at every step, we can maintain a balanced curriculum that avoids training with problems that are either too trivial or too challenging. In parallel, an online difficulty filter discards trivial rollouts—such as those that the model always fails or always solves—ensuring that we only train on rollouts with meaningful learning signal. Together, these mechanisms allow the curriculum to evolve as the model's capabilities increase, and continue to provide good learning signal.

\subsubsection{Scaling Sequence Length}

Over the course of RL training, the sequence length of model generations naturally increases~\cite{deepseekai2025deepseekr1}. Training on these increasingly long rollouts while preserving efficiency is nontrivial. In our training setting, we managed to reach up to \(48\mathrm{k}\) sequence length with FSDP degree equal to 32, by leveraging aggressive activation check-pointing and Flash Attention 3 (FA3) ~\cite{shah2024flashattention}. However, for some of the more difficult environments, sequence length of at least \(64\mathrm{k}\) was required. To solve this, we explored the following two approaches:

\textbf{Context Parallelism.} With increasing sequence length, the attention score matrix becomes the dominant component of the training memory footprint. Algorithms such as \texttt{FlashAttention}~\cite{dao2022flashattentionfastmemoryefficientexact} help mitigate this, but are often not sufficient. 
Context parallelism distributes the attention computation across \(N_{\mathrm{cp}}\) GPUs, reducing the memory footprint at the cost of communication overhead. 
The most common implementation is based on \emph{Ring Attention}~\cite{liu2023ringattentionblockwisetransformers}, which assigns separate \texttt{chunks of Q}, \texttt{K}, and \texttt{V} to each GPU and rotates \texttt{K} and \texttt{V} among devices to compute the attention matrix. 
This algorithm is implemented in PyTorch; however, its implementation for \texttt{FlexAttention}~\cite{dong2024flexattentionprogrammingmodel} was experimental at the time of our training. 
Although we were able to scale the sequence lengths up to \(256\mathrm{k}\) using this approach with \(N_{\mathrm{cp}} = 2\), doing so effectively halved our data-parallelism degree and additionally exhibited accuracy degradations, making it unsuitable for our production training setting.

\textbf{Activation Offloading.} Our training utilized full activation checkpointing, meaning that only the outputs of each decoder layer and the top-level module activations are stored, while all intermediate activations are recomputed during the backward pass. Ignoring top-level activations, the activation memory for a sequence length of 
\(48\mathrm{k}\), hidden size \(4096\), and \(46\) decoder layers is

\[
\text{Mem}_{\mathrm{act}}
= 46 \times (48{,}000 \times 4{,}096) \times 2~\text{bytes}
\approx 18~\text{GB}.
\]

To reduce this footprint, we offload activations to the CPU using an implementation based on \texttt{torchtune}~\cite{torchtune}. This enabled us to scale the sequence length to \(72\mathrm{k}\) with the same hardware configuration as above, without any degradation in MFU. We observed a memory leak in the asynchronous offloading implementation when using CUDA streams, and therefore adopted the synchronous variant. Although this led to a slight decrease in MFU, the impact was negligible at roughly \(0.1\%\).

\subsubsection{Distributed Muon} 
\label{subsubsec:distributed-muon}

As shown by ~\cite{liu2025muonscalablellmtraining}, reusing the Muon optimizer during post-training yields the best performance when the model was initially pretrained with Muon. Unlike SGD or Adam, which apply element-wise updates, Muon~\cite{jordan2024muon} operates at the matrix level: its Newton-Schulz update requires access to the full gradient tensor. Consequently, we cannot directly apply Muon to FSDP-sharded gradients. Gathering full gradients on every rank and redundantly performing the same Muon computation would be prohibitively expensive.

To address this, we explore two strategies for distributing Muon across multiple nodes. Our first approach uses an overlapping round-robin scheme: each rank gathers a subset of FSDP-sharded gradients based on its index, computes the Newton-Schulz update locally, and then scatters the updated gradients back to all FSDP ranks. This approach parallelizes the expensive computation across ranks and can hide the additional communication. However, at large multi-node scales, issuing many overlapping gathers leads to InfiniBand congestion.

We therefore adopt a more efficient method based on all-to-all collectives, which reshuffles gradient shards without relying on many individual gathers. Although this design is less flexible and may require padding tensors before communication, it significantly improves performance and avoids congestion at scale. We use this all-to-all–based algorithm for our main training runs, leveraging the open-source implementation provided in Dion~\cite{ahn2025dion}.
 
\subsubsection{Efficient Mixture-of-Experts Support}
\label{subsec:moe}

\begin{figure}[ht]
\centering
\includegraphics[width=1.0\textwidth]{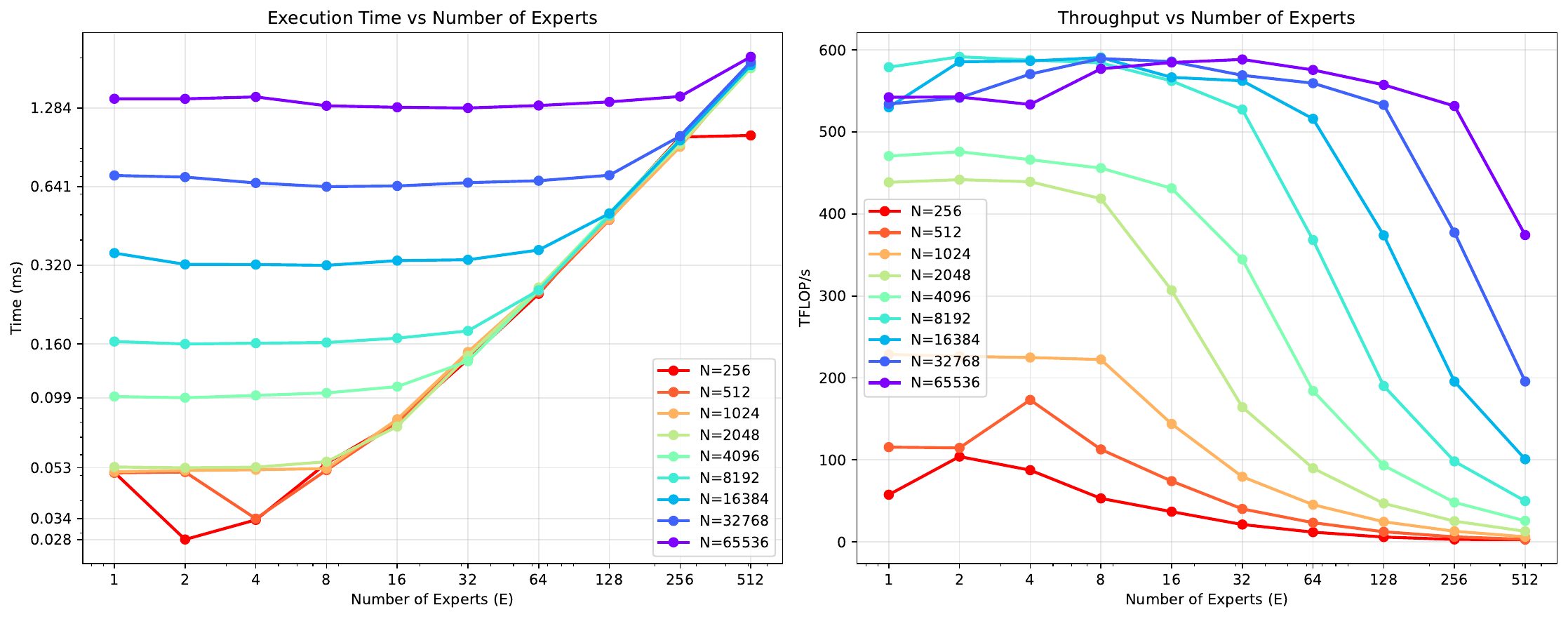}
\captionsetup{width=1.0\textwidth}
\caption{Execution time and TFLOPS of \texttt{torch.\_grouped\_mm} with hidden dim $4096$ and MoE dim $1408$ on H200 SXM at different sequence lengths and number of experts. We assume that the input is perfectly balanced between the experts and thus an increase in experts leads to an inversely proportional decrease in the number of tokens and work per expert, eventually causing lower TFLOPS as the work per expert is no longer able to saturate the kernel. At sequence lengths (N) $32,768$ and $65,536$, the TFLOPS remains in the saturated regime up to $128$ experts. We thus do not gain significant throughput from using expert parallel given our training parameters.}
\label{grouped_mm_bench}
\end{figure}

We use the Mixture-of-Experts (MoE)~\cite{shazeer2017moe} layer implementation from \torchtitan~\cite{liang2025torchtitan} to leverage the efficient grouped matrix multiplication kernels for expert execution which comes with support for expert parallelism (EP).
We found that enabling EP led to worse training throughput and thus did not enable it for our training.
The reduced throughput can be attributed to the relatively large sequence length and hidden dimension on each GPU for our training runs.

We train with relatively large sequence length and hidden dimension and are thus in the regime depicted in Figure~\ref{grouped_mm_bench} where we already saturate the grouped gemm kernel without the need for decreasing the number of experts per GPU with expert parallel.
Expert parallel in this regime will lead to increased overhead from the scatter and gather without decreasing the time spent performing the grouped gemm.
Training that uses lower sequence length, hidden dimension or utilizes parallelizations that decrease the amount of work per GPU like context parallel and tensor parallel might see an improvement from using expert parallelism.

To maintain compatibility with HF, as for example required by our vLLM inference engine, we transform the state dict on-the-fly during the broadcasting from the \torchtitan-based MoE layer implementation used by the trainer to the HF-based MoE layer implementation used by the inference engine.

To monitor expert load distribution, we compute and log the maximum violation load-balancing metric $\mathrm{MaxViolation} = \frac{\max_i \mathrm{Load}_i - \overline{\mathrm{Load}_i}}{\overline{\mathrm{Load}_i}}$ as described in \citep{wang2024auxiliarylossfreeloadbalancingstrategy}.
This metric quantifies the degree to which expert load imbalance slows down the MoE layer relative to an ideally balanced configuration.
Since imbalance directly affects both training and inference throughput, it provides a useful diagnostic of system efficiency.

\subsection{Verifiers: Environments for LLM Reinforcement Learning}
\label{subsec:verifiers}

We train \INTELLECT ~using environments built with our \verifiers ~library, which provides a set of modular and extensible components for concisely expressing complex environment logic while maintaining highly scalable performance. Conceptually, \verifiers\ environments for RL training play the same role as datasets for SFT or pre-training: disentangling environments from training infrastructure yields desirable compositionality and interoperability, allowing environments to be developed, tested, and versioned independently of the trainer.

\subsubsection{Environment Design}

Environments built with \verifiers ~are installable Python modules, and consist of:
\begin{itemize}
    \item a \texttt{dataset} where each row corresponds to a task example, including the input prompt and any necessary metadata for execution scoring (e.g.\ ground-truth answer, test cases);
    \item a \texttt{rollout} method which takes as input a dataset row and an OpenAI-compatible inference client, executes all steps of action until a termination condition is reached, and collects all information required for training (e.g. token ids, logprobs);
    \item a \texttt{Rubric} object which includes one or more reward functions (operating either on a per-rollout or per-group basis), and logs final scores and metrics upon completion of rollouts;
    \item a \texttt{load\_environment} method implemented in the environment module which instantiates the environment, handling any necessary preprocessing and resource provisioning.
\end{itemize}

\paragraph{Rollout Orchestration.}

Rollouts are executed asynchronously via \texttt{asyncio}, allowing thousands of concurrent rollouts to proceed in parallel. Inference requests, tool calls, and reward functions are dispatched and awaited independently of other in-flight rollouts. Parallelism occurs at multiple granularities: we replicate across inference workers, API clients, and environment processes via a central orchestrator. We apply fine-grained semaphore-based throttling to ensure inference workers are kept busy while minimizing KV cache eviction.

\paragraph{Rubrics and Reward Functions.}

The \texttt{Rubric} abstraction manages reward computation with support for multiple weighted reward functions. Each reward function receives the prompt, completion, ground-truth answer, and rollout state, and returns a scalar score. Scores from multiple reward functions are combined via configurable weights to produce a final reward signal.

For more complex scoring scenarios, rubrics can be composed to aggregate multiple scoring strategies (e.g.\ combining a format-checking rubric with an LLM judge rubric). The scoring interface can also be overridden to implement inter-group comparisons such as voting, ranking, or relative scoring across samples from the same problem.

\paragraph{Evaluations.} We additionally use \verifiers ~environments directly for evaluation, both offline (via a standalone CLI which runs remote-hosted evaluations via the Environments Hub) and online during training. The same \texttt{rollout} and \texttt{Rubric} entrypoints can be used for either training or evaluation, ensuring consistency across deployment settings.

\paragraph{Extensibility via Class Inheritance.}

The environment hierarchy provides progressive specialization for common use cases. The base class handles dataset management, prompt formatting, and the core generate/score pipeline. Multi-turn environments extend this with a rollout loop that alternates between model responses and environment responses until a termination condition is met. Single-turn environments provide a minimal specialization for tasks requiring only a single model response. Tool-calling environments further specialize multi-turn behavior with native OpenAI-format tool calling: tool definitions are automatically converted to the API schema, tool calls in model responses are parsed and executed, and results are appended as tool messages. Custom environments inherit from the appropriate base class and override methods to implement task-specific logic such as termination conditions and environment response generation.

\begin{figure}[h]
\centering
\begin{tikzpicture}[
    node distance=1cm,
    box/.style={draw, rounded corners, minimum width=2.8cm, minimum height=0.6cm, font=\small\ttfamily},
    arrow/.style={->, thick}
]
\node[box] (python) {CodeEnv};
\node[box, below of=python] (sandbox) {SandboxEnv};
\node[box, below of=sandbox] (stateful) {StatefulToolEnv};
\node[box, below of=stateful] (tool) {ToolEnv};
\node[box, below of=tool] (multi) {MultiTurnEnv};
\node[box, below of=multi] (base) {Environment};

\draw[arrow] (python) -- (sandbox);
\draw[arrow] (sandbox) -- (stateful);
\draw[arrow] (stateful) -- (tool);
\draw[arrow] (tool) -- (multi);
\draw[arrow] (multi) -- (base);
\end{tikzpicture}
\caption{\textbf{Inheritance Hierarchy. } Class hierarchy for \texttt{CodeEnv}, used in the \texttt{primeintellect/i3-code} environment. Each level adds functionality: \texttt{Environment} provides the core abstraction; \texttt{MultiTurnEnv} adds iterative rollout logic; \texttt{ToolEnv} adds OpenAI-format tool calling; \texttt{StatefulToolEnv} enables injecting tool arguments that depend on rollout state (e.g.\ resource IDs); \texttt{SandboxEnv} manages containerized execution environments; and \texttt{CodeEnv} runs test cases against the code generated by the LLM.}
\label{fig:env-hierarchy}
\end{figure}
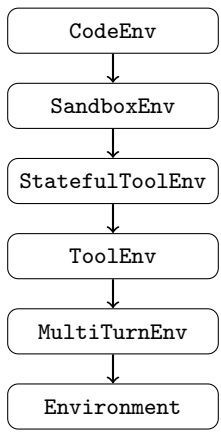

\subsubsection{Integration with \primerl}
\primerl\ has first-class support for \verifiers\ environments, which are installed as standalone Python modules via the \envhub\footnote{\url{https://hub.primeintellect.ai}}. Environments can be developed and tested in isolation against local or API models, then pushed to the \envhub\ and used immediately in training without any code modification in \primerl. The orchestrator loads environments by their Python module identifiers, invokes them with batches of inputs and inference clients, and receives finished rollout states, including rewards, token IDs, logprobs (directly from vLLM), and attention masks.

\paragraph{Multi-Environment RL Training.}

The \texttt{EnvGroup} pattern in \verifiers ~allows the combination of multiple environments into a single class object with concatenated datasets, where an injected task ID column is used to route rollout and scoring logic across appropriate sub-environments. We leverage this functionality in \primerl\ to support simultaneous training of \INTELLECT ~across many environments without needing any explicit multi-environment-aware code within the orchestrator after the \texttt{EnvGroup} is instantiated.

\subsubsection{Environments Hub}

Many RL frameworks treat environments as subfolders within a central training repository, tightly coupling environment logic to training infrastructure. This complicates versioning, makes it difficult to run controlled ablations across environment variants, and creates friction for external contributors who must navigate the full training codebase to add or modify tasks or run offline evaluations.

The \envhub~addresses these issues by providing an open registry where environments built with \verifiers\ are packaged as standalone Python modules with pinnable dependencies and a standardized entry point. Environments can be versioned, shared, tested, and iterated on outside of the context of training code, enabling researchers to contribute new tasks independently. Combined with our open-source \primerl~trainer, sandboxes for secure code execution, and distributed compute infrastructure, the \envhub\ forms part of a full infrastructure stack for open RL research, lowering the barrier for anyone to train, evaluate, and fine-tune models on custom tasks.

\subsubsection{Evaluations}

Evaluation and training are tightly coupled through the use of \verifiers, streamlining the process of evaluating against a wide range of common benchmarks. Evaluation can be run both as part of an active training (online) or as a standalone entrypoint (offline). When evaluating online, the orchestrator asynchronously interleaves evaluation requests with training requests using the same inference pool as the trainer, effectively hiding evaluation overhead while providing real-time feedback of training performance.

\subsection{Prime Sandboxes: Code Execution for RL Training}
Executing untrusted code for thousands of concurrent rollouts requires a container orchestration layer capable of sub-second provisioning and millisecond-level execution latency. While Kubernetes provides the primitives for container management, standard architectural patterns are insufficient for the throughput required by high-velocity training.
\subsubsection{The Limits of Naive Orchestration}
A standard, naive implementation of a remote execution sandbox typically relies on the Kubernetes API Server. In this design, the training loop utilizes standard client libraries to spawn ephemeral pods and executes commands via \texttt{kubectl exec}. This operation relies on upgrading an HTTP request to a WebSocket connection, which is then proxied through the Kubelet to the container runtime.
Empirical testing revealed that this approach is fundamentally unscalable. While the code execution itself should take milliseconds, the orchestration overhead creates latencies measured in seconds. Because every execution command involves an authenticated API request that is logged and persisted in \texttt{etcd}, the control plane becomes the primary bottleneck. Etcd is Kubernetes' distributed key-value store that maintains cluster state; every API operation must be serialized through write locks, creating a fundamental throughput ceiling. At a scale of thousands of concurrent sandboxes, we measured execution latency spiking to 2.5 seconds per command due to API server saturation and \texttt{etcd} write-lock contention.
\subsubsection{Prime Sandboxes Architecture}
To overcome these control plane limitations, we developed \textit{Prime Sandboxes}, which is a cloud based, Sandbox execution infrastructure which covers the requirements for agentic RL at scale.
We re-engineered the communication path to bypass the Kubernetes API for the critical execution loop.
We introduced a high-performance Rust-based Gateway that accepts execution requests via a lightweight HTTP API. Instead of routing through the Kubernetes control plane, this Gateway communicates directly with the sandbox pods via Kubernetes Headless Services. A Headless Service allows the Gateway to resolve the direct IP addresses of individual pods via DNS, bypassing the overhead of \texttt{kube-proxy} load balancing.

However, relying on DNS for thousands of ephemeral pods creates a secondary scalability challenge. Standard CoreDNS configurations can become overwhelmed by the high churn rate of A-records associated with headless services, leading to resolution throttling. To ensure the Gateway can resolve pod IPs in milliseconds, we deployed a custom, high-throughput CoreDNS architecture optimized for high-velocity record updates.

The sandbox pod itself utilizes a Sidecar pattern. A privileged sidecar container functions as an execution agent. Upon receiving a request from the Gateway, the sidecar utilizes \texttt{nsenter} to inject commands directly into the target's namespace. This architecture achieves the speed of a local process spawn while maintaining full container isolation.
\subsubsection{Asynchronous Lifecycle Management}
Reliably detecting when a sandbox is ready for execution is a critical latency path. The naive approach involves either actively polling the Kubernetes API to check for the \texttt{Running} state or relying on a standard Controller to watch the event stream.
While sufficient for low-volume workloads, this method fails under high concurrency. Polling thousands of pods creates immense read pressure on the API server. Furthermore, standard controllers—specifically those built with Kopf (Kubernetes Operator Pythonic Framework)—typically process reconciliation queues sequentially. In a "thundering herd" scenario where thousands of pods initialize simultaneously, the resulting notification backlog causes the system to report a sandbox as "Ready" seconds after it has actually booted, leaving valuable compute resources idle.

To eliminate this latency, we inverted the status reporting flow. We utilize Kopf strictly for asynchronous maintenance tasks, such as error handling and resource finalization. For readiness signaling, we bypass the Kubernetes control plane entirely: the sandbox sidecar transmits a direct webhook to the training backend the moment it becomes operational. This "push-based" architecture ensures the training loop is notified within milliseconds of boot completion, achieving a consistent end-to-end cold start—from request to operational sandbox with arbitrary user-specified container images—of under 10 seconds regardless of cluster load. For pre-warmed environments using standard runtime images, acquisition is effectively instantaneous.

\subsubsection{Image Distribution and Infrastructure Density}
Distributing container images to thousands of sandboxes simultaneously presents a distinct scalability barrier. A naive implementation that pulls images from public repositories (e.g., Docker Hub) hits fatal bottlenecks immediately: upstream registries enforce strict rate limits that block high-concurrency requests, and the sheer volume of data transfer saturates node network bandwidth, delaying boot times by minutes.
To overcome these physical constraints, we architected a two-tiered image distribution strategy:
\begin{itemize}
\item \textbf{Custom Registry and Image Streaming:} For dynamic environments requiring unique dependencies, we host a private, high-throughput container registry. Crucially, we utilize Container Image Streaming (Lazy Pulling). Instead of blocking startup until the entire image manifest is downloaded, the runtime fetches only the data chunks necessary for the entrypoint process. The remaining layers are streamed in the background, allowing the sandbox to become operational seconds before the full image is physically present on the node.
\item \textbf{Warm Pools:} For environments relying on static runtime images (e.g., standard Python distributions), we maintain a "warm pool" of pre-provisioned pods. This allows the training loop to acquire a sandbox instantly without incurring any image pull latency or initialization overhead.
\end{itemize}
Underlying this distribution layer is a custom Cluster Autoscaler and bin-packing scheduler designed for extreme density. We target a packing factor of 256 sandboxes per node and utilize the \textit{Burstable} QoS class. Burstable QoS allows pods to request a baseline of CPU resources but burst above that limit when available, which is ideal for RL workloads where sandboxes alternate between brief execution spikes and idle waiting periods. This allows the cluster to vastly oversubscribe CPU resources during the idle periods inherent in RL training steps, maximizing hardware efficiency without compromising peak execution throughput. 

\subsubsection{Security and Capabilities}
We utilize gVisor (runsc) as the container runtime, providing a user-space kernel that isolates the host from potential exploits within the untrusted code as well as configurable network policies to limit the communication allowed from within a sandbox. Beyond simple code execution, the architecture supports complex network requirements, allowing sandboxes to expose arbitrary TCP/UDP ports for HTTP traffic. Furthermore, the system is designed to support hardware acceleration, enabling the mounting of GPUs for environments dependent on custom GPU kernels.

\subsection{Compute Orchestration: Frontier GPU Infrastructure}

We deployed 512 NVIDIA H200 GPUs across 64 interconnected nodes. The primary engineering challenge lies in maintaining determinism and synchronization across a distributed system prone to hardware failures.

\paragraph{Provisioning and Fabric.}
To eliminate configuration drift, we enforce a strict Infrastructure as Code paradigm using idempotent Ansible playbooks that handle dynamic hardware discovery and automated firewall generation. Distributed training performance is bound by the tail latency of the \textit{AllReduce} collective, so we utilize a 400Gbps NDR InfiniBand fabric (NVIDIA ConnectX-7) and validate throughput before every run, targeting $\geq$160 GB/s. When performance degrades, an automated binary search isolates straggler nodes with faulty transceivers.

\paragraph{Orchestration.}
We use Slurm with Cgroup v2 integration to guarantee resource reclamation---upon job termination, the kernel freezes and eliminates the entire cgroup hierarchy, preventing zombie processes from holding GPU memory. This provides container-like isolation without filesystem overhead.

\paragraph{Storage.}
A tiered architecture balances I/O performance: Lustre handles high-throughput operations (training trajectories, multi-terabyte checkpoints), while NVMe-backed NFS serves metadata-heavy user environments and enables keyless SSH across the fleet.

\paragraph{Observability.}
We monitor GPU telemetry via DCGM aggregated into Prometheus, with active alerting on Xid errors and thermal slowdown events. This allows proactive node draining before a failing component corrupts training progress. 

\section{\INTELLECT\ Training}
\label{sec:intellect-3-training}

We train \INTELLECT\ in two main stages: a supervised fine-tuning stage  and a large-scale RL stage. We use GLM-4.5-Air base as our base model. Both stages, including multiple ablations, were carried out on a 512 H200 cluster over the course of two months. For the entire stack, from the RL environments to the evaluation of the model, we utilize open-source environments contributed to our \envhub.

\subsection{Environments Mix}

We train on a diverse and challenging mix of environments designed to enhance the reasoning and agentic capabilities of our model.

\subsubsection{Math}

For our model to excel in mathematical problem solving, we carefully design our math environment \footnote{\pilogo ~\href{https://app.primeintellect.ai/dashboard/environments/primeintellect/i3-math}{\texttt{environments/primeintellect/i3-math}}} for long chain-of-thought reasoning.
It consists of 21.2K challenging math problems, curated from Skywork-OR1~\cite{he2025skyworkor1}, Acereason-Math~\cite{chen2025acereasonnemotronadvancingmathcode}, DAPO~\cite{yu2025dapoopensourcellmreinforcement}, and ORZ-Hard~\cite{hu2025openreasonerzeroopensourceapproach}.
To verify the model's responses we parse out the final answer and compare it against the ground truth using \texttt{math-verify}~\cite{kydlicek2025mathverify}. In practice, we have found a non-negligible fraction of false negatives with mere rule-based verification. For this reason, we additionally employ \texttt{opencompass/CompassVerifier-7B}~\cite{CompassVerifier} as an LLM-judge to double-check all answers which are marked as wrong by the rule-based verifier.
Finally, we difficulty annotate the entire dataset by computing the average solve rate of \texttt{Qwen/Qwen3-4B-Thinking-2507}~\cite{qwen3technicalreport} over eight generations per problem. We use these annotations to filter our samples which are too easy at various stages of our post-training pipeline.
    
\subsubsection{Code} Our code environment\footnote{\pilogo  ~\href{https://app.primeintellect.ai/dashboard/environments/primeintellect/i3-code}{\texttt{environments/primeintellect/i3-code}}} tasks the model with single-turn programming challenges in Python. The environment takes inspiration from DeepCoder \cite{deepcoder2025} and makes heavy use of our \texttt{SYNTHETIC-2} dataset \cite{primeintellectteam2025synthetic2}. Solutions are verified by executing up to 15 test cases per problem inside \textit{Prime Sandboxes}. During training, asynchronous and isolated execution per solution is facilitated by over 4000 concurrent sandboxes. On any sandbox failure, we mask out the corresponding model completion. For difficulty filtering, we annotate the 8.6K examples with the average solve rate of \texttt{Qwen/Qwen3-4B-Instruct-2507}~\cite{qwen3technicalreport} over eight generations per problem.

\subsubsection{Science} 

We use the science environment\footnote{\pilogo ~\href{https://app.primeintellect.ai/dashboard/environments/primeintellect/i3-science}{\texttt{environments/primeintellect/i3-science}}} to improve our model's capabilities in domains such as physics, chemistry, and biology. Similarly to our math environment, we use both math-verify and an LLM-judge to verify the answers. The dataset we use consists of 29.3K challenging problems spanning diverse domains curated and filtered from MegaScience~\cite{fan2025megasciencepushingfrontiersposttraining}.
Again, we annotate the dataset for difficulty by computing the average solve rate of \texttt{Qwen/Qwen3-4B-Instruct-2507}~\cite{qwen3technicalreport} over 16 generations per problem, which we use to filter out too easy samples at various stages of the post-training pipeline.

\subsubsection{Logic}

Our logic environment\footnote{\pilogo ~\href{https://app.primeintellect.ai/dashboard/environments/primeintellect/i3-logic}{\texttt{environments/primeintellect/i3-logic}}} includes a diverse set of 29 logical tasks, puzzles and games, such as evaluating boolean expressions, solving cross-word puzzles or Sudoku, or playing Minesweeper. Both the 11.6K problems and verifiers were adapted from SynLogic~\cite{liu2025synlogicsynthesizingverifiablereasoning}. Similarly to our other environments, we compute the solve rate of \texttt{Qwen/Qwen3-4B-Instruct-2507}~\cite{qwen3technicalreport} over 16 trials to gauge the difficulty of the problems, in order to selectively filter out samples during post-training.

\subsubsection{Deep Research}

Our web search environment\footnote{\pilogo ~\href{https://app.primeintellect.ai/dashboard/environments/primeintellect/deepdive}{\texttt{environments/primeintellect/deepdive}}}~\cite{lu2025deepdiveadvancingdeepsearch} provides the model with a search tool that uses Serper~\cite{serper2025} to return an enumerated list of search results, a click tool to retrieve the markdown-formatted textual content of webpage chosen by its index in the previous search results, an open tool that returns a website's content in the same way as the click tool, given a URL, and a finish tool which the model uses to provide its final answer. The environment tasks the model with answering questions from the dataset using the given tools, and rewards it with 1 for a correct answer and 0 for an incorrect one. Optionally, a redundancy penalty can be applied to the queries of subsequent searches, which we set to zero. We leverage a corpus of training examples from z-AI's DeepDive dataset, which consists of complex, multi-step questions extracted from open knowledge graphs with the help of LLMs. It contains 1K samples for SFT trajectory generation and 2.2K samples for RL.

\begin{figure}[ht]
\centering
\includegraphics[width=0.6\textwidth]{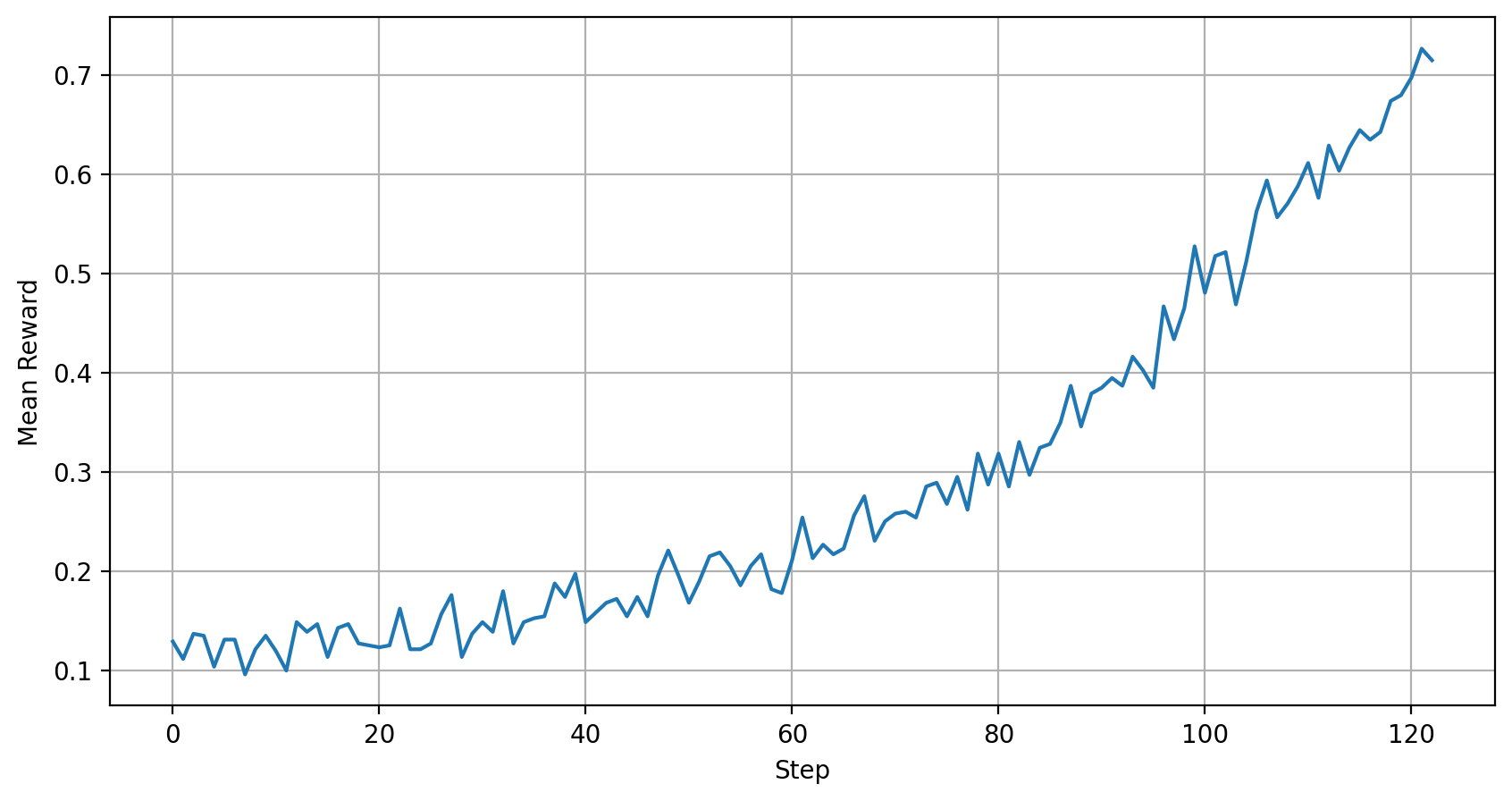}
\captionsetup{width=0.7\textwidth}
\caption{Mean reward of \texttt{Qwen/Qwen3-4B-Instruct-2507}~\cite{qwen3technicalreport} over RL training steps on DeepDive after a short SFT phase.}
\label{deepdive-reward}
\end{figure}

To validate the implementation of our environment, we trained \texttt{Qwen/Qwen3-4B-Instruct-2507}~\cite{qwen3technicalreport} using SFT on the public DeepDive traces for 26 steps at batch size 34, for a total of 884 samples, followed by 122 steps of RL at a group size of 16 and a total batch size of 512. Figure \ref{deepdive-reward} shows the mean reward per step over the course of RL training. The success of a 4 billion parameter model to learn Deep Research via RL on DeepDive demonstrates the correctness and usability of our DeepDive environment implementation.

\subsubsection{Software Engineering}

We developed two Software Engineering (SWE) environments \footnote{\pilogo  ~\href{https://app.primeintellect.ai/dashboard/environments/primeintellect/deepswe}{\texttt{environments/primeintellect/deepswe}}}\footnote{\pilogo  ~\href{https://app.primeintellect.ai/dashboard/environments/primeintellect/mini-swe-agent-plus}{\texttt{environments/primeintellect/mini-swe-agent-plus}}}. They implement two modified agent scaffolds, R2E-Gym \cite{jain2025r2e, deepswe2025} and mini-swe-agent-plus \cite{wang2025klearagentforgeforgingagenticintelligence}. Further, they include three sandbox harnesses supporting common formats for SWE datasets, images and test suites like R2E-Gym \cite{jain2025r2e}, SWE-smith \cite{yang2025swesmith} and Multi-SWE-bench \cite{zan2025multiswebench}.
For the R2E-Gym scaffold, we swapped out the \texttt{finish()} for the \texttt{submit()} tool as the \texttt{result} parameter serves no use internally and is being called verbosely by the model. We adapt the mini-swe-agent-plus scaffold for native reasoning and tool use by changing its prompt and replacing code block parsing with tool calling.
Inside a sandbox the agent can navigate the repository of a given Github project and is tasked with fixing an issue. The scaffold equips the model with tools for executing Bash commands and editing files. The maximum number of turns the agent can take is capped at 200. After submitting the solution, the test suite of the repository runs to determine whether the correct tests change their status from failing to passing. As with our code environment, when a given sandbox fails we mask out the model's completion and cancel the generation.
To render ephemeral and nearly instant rollouts possible we deploy our Custom Registry with \textit{Prime Sandboxes} hosting over 20,000 images containing pre-installed Github repositories \footnote{\pilogo  ~\href{https://docs.primeintellect.ai/sandboxes/overview}{\texttt{Prime Sandboxes}}}.

\subsection{Supervised Fine-Tuning}

Prior to RL, we run two complementary supervised fine-tuning (SFT) phases. First, we conduct a large-scale general chat-and-reasoning SFT stage that strengthens the model’s conversational abilities and core reasoning skills. Second, we run an agentic SFT stage focused on improving the model’s ability to use tools effectively and operate within long-horizon, agentic workflows. Together, these stages establish a strong prior and a stable behavioral foundation for the subsequent RL phase.

\textbf{General Reasoning SFT.} For our first SFT stage, we construct a large-scale dataset spanning diverse domains and leverage many high-quality, permissive open-source datasets. Our two main sources are the math, code, science, and tool splits from NVIDIA's \texttt{Nemotron-Post-Training-Dataset-v1}~\cite{NemotronPostTrainingDatasetV1,bercovich2025llamanemotronefficientreasoningmodels} and the chat and instruction following splits from AM's \texttt{AM-DeepSeek-R1-0528-Distilled}~\cite{AM-DeepSeek-R1-0528-Distilled} dataset. Both contain synthetically generated reasoning traces from \texttt{DeepSeek-R1-0528}.  During training we respect the natural ratios of the datasets. We train a full epoch with $\sim$33M tokens per step at context length 65K. We use the Muon optimizer with weight decay 0.01 and learning rate 5e-5, warmed up linearly from 1e-8 over 300 steps. We rely on FSDP with a world size of 64 to efficiently shard the model and DP replicate size 8, spanning training across the full cluster of 512 GPUs

\begin{table}[t]
\centering
\caption{\textbf{SFT Data Sources}}
\begin{tabular}{lcccc}
\toprule
\textbf{Dataset} & Num. Examples & Num. Tokens & Stage 1 & Stage 2 \\
\midrule
OpenReasoning-Math & 2M & 78.1B & \checkmark & \checkmark \\
OpenReasoning-Code & 1.9M & 94.3B & \checkmark & \checkmark \\
OpenReasoning-Science & 310K & 32B & \checkmark & \checkmark \\
OpenReasoning-Tool & 800K & 3.8B & \checkmark & \checkmark \\
AM General Chat & 952K & 8.4B & \checkmark & \checkmark \\
AM Instruction Following & 54K & 400M & \checkmark & \checkmark \\
\midrule
SWE Swiss & 10.3K & 700M & & \checkmark \\
Toucan Tool & 116K & 700M & & \checkmark \\
Environments Mix & 38.4K & 1.9B & & \checkmark \\ 
\bottomrule
\end{tabular}
\end{table}

\textbf{Agentic SFT.} Following the general chat-and-reasoning SFT phase, we conduct a second supervised fine-tuning stage targeted at agentic behavior, tool use, and long-horizon control. Whereas the first SFT phase focuses on conversational competence and long chain-of-thought reasoning in STEM domains, the agentic SFT phase is smaller, and curated to endow the model with robust capabilities for calling external tools, maintaining coherent state over long-running tasks, and operating effectively in extended sequences. To this end, we combine multiple open-source agentic datasets such as SWE-Swiss~\cite{SWESwiss2025}, and Toucan Tool~\cite{xu2025toucan}, as well as synthetically-generated datasets created from other environments on the \envhub\ using \texttt{DeepSeek-R1-0528}. All datasets were processed to ensure consistent tool call formatting, filtered for English content, and standardized so that they are compatible with our trainer.

This stage also serves a complementary purpose: pushing the model toward longer effective context lengths, ensuring stability and competence beyond 65K context window. We leverage context parallelism (CP) to effectively scale to 98K context length training, ensuring that our model learns to maintain consistency in long-running agentic tasks. We train for two epochs, resuming from our final stage 1 checkpoint. Again, we use the Muon optimizer starting with a learning rate of 5e-8, and decay it linearly over the full 800 steps of training.

\begin{figure}[ht]
\centering
\begin{subfigure}[c]{0.49\textwidth}
\centering
\includegraphics[width=\textwidth]{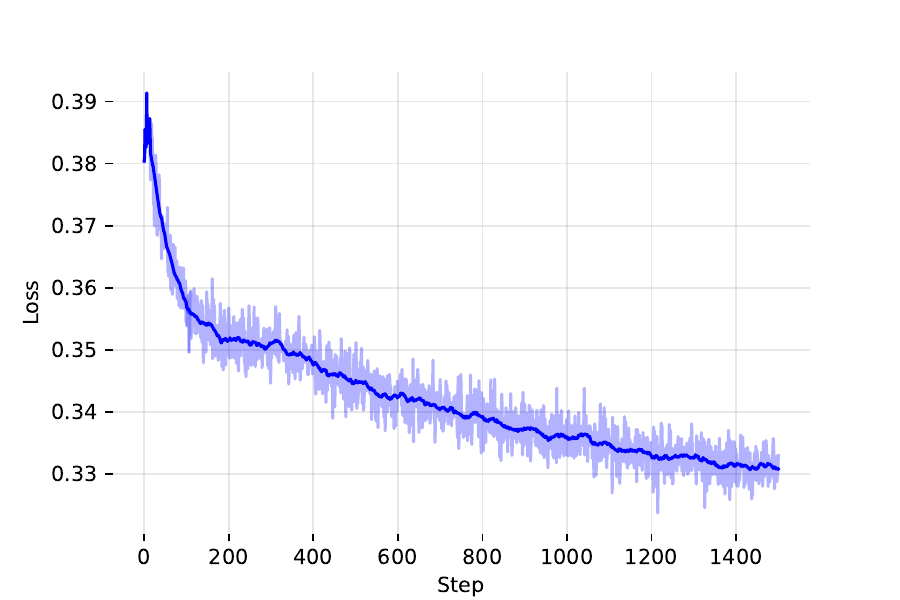}
\caption{Stage 1}
\label{fig:sft-stage-1}
\end{subfigure}
\hfill
\begin{subfigure}[c]{0.49\textwidth}
\centering
\includegraphics[width=\textwidth]{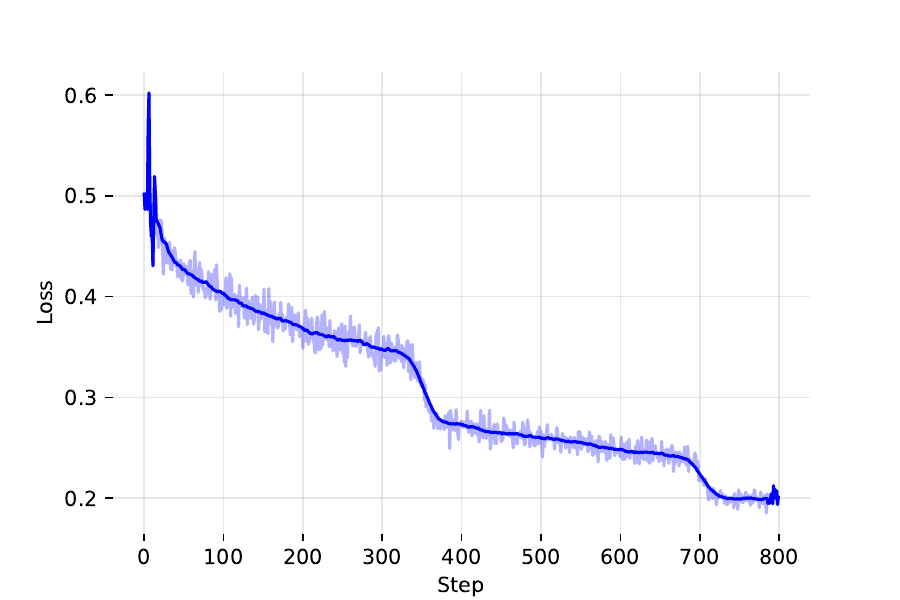}
\caption{Stage 2}
\label{fig:sft-stage-2}
\end{subfigure}
\caption{\centering \textbf{Supervised Fine-Tuning.} We show the loss curves of our general reasoning SFT stage (left) and the agentic SFT stage (right) over the course of training. Both runs show smooth optimization without any loss spikes.}
\end{figure}

\textbf{Chat Template.} The \INTELLECT\ chat template is inspired by Qwen3- and GLM-family of models. The template uses familiar control tokens—such as \texttt{<|system|>}, \texttt{<|user|>}, and \texttt{<|assistant|>}—to mark roles in multi-turn conversations, and \texttt{<|im\_start|>} and \texttt{<|im\_end|>} to delimit conversational turns. Tool calls follow XML-style tagging format.

We introduce the following modifications: First, our model always reasons, with no user-exposed reasoning-effort controls. This is implicitly baked into the model by dominantly training on reasoning-only SFT traces, and explicitly by setting appending a \texttt{<|think|>} token via the chat template. To ensure proper usage of our model, use the \texttt{qwen3\_coder} tool call parser, and the \texttt{deepseek\_r1} reasoning parser. To retain thinking across turns, the chat template automatically parses \texttt{reasoning\_content} field to ensure reasoning chains are consistently represented without requiring manual formatting.

\subsection{Reinforcement Learning}

For the RL portion of the training, we used a batch size of 256 prompts with 16 rollouts per prompt and a maximum context length of 65536.
We utilize online difficulty filtering, as well as an easy difficulty pool to remove any prompt with a pass rate of 1 from being sampled again, as these prompts would contribute no learning signal.
Our \texttt{max\_off\_policy\_steps} is set to 8 to ensure we remove any excessively off-policy rollout.
We use Muon with a learning rate of $1e-6$.
During training, the data mix is carefully tuned to balance performance across domains.

% training / inference worker ratio
We use 60 nodes in total for our RL training, each with 8 H200s.
We allocate nodes between training and inference at about a 1:3 ratio, with 16 nodes for training and 44 nodes for inference to get the best throughput.
% step time
We observe a step time of $\sim 1500s$ per step when training at $65,536$ sequence length with in-flight weight updating.
Without in-flight weight updating, we observe an increase in step time of more than $2\times$ as the inference is significantly less efficient.

\label{eq:icepop}
\textbf{Training Algorithm. } We adopt masked token-level importance sampling~\cite{icepop2025}. For a batch of $N$ rollouts, we use the formulation below.

\begin{align}
\label{eq:icepop}
\mathcal{J}_{{\text{IcePop}}}(\theta) &= 
\mathbb{E}_{
x \sim \mathcal{D}, 
\{y_i\}_{i=1}^N \sim \pi_{\textcolor{red}{\text{infer}}}}
\left[ \frac{1}{\sum_{i=1}^N |y_i|}\sum_{i=1}^N
\sum_{t=1}^{|y_i|}
\Big[\mathcal{M}\Bigl(
\frac{\pi_{\textcolor{blue}{\text{train}}}(y_{i,t} \mid x, y_{i,<t};\theta)}{\pi_{\textcolor{red}{\text{infer}}}(y_{i,t} \mid x, y_{i,<t}; \theta_{\mathrm{old}})};\alpha, \beta
\Bigr) \right. \widehat{A}_{i,t} \Bigg]
&
\end{align}

\begin{equation}
\label{eq:icepop-masking}
\mathcal{M}(k) =\begin{cases} k & \text{if \ } k \in [\alpha, \beta] \\ 0 & \text{otherwise}\end{cases}
\end{equation}

where $\pi_{\textcolor{red}{\text{infer}}}$ refers to the policy that generated the rollout, $\pi_{\textcolor{blue}{\text{train}}}$  refers to the current trainer policy, $\mathcal{M}(\cdot; \alpha, \beta)$ is the masking function from Eq.~\ref{eq:icepop-masking}. The token-level advantage is estimated as $\hat{A}_{i,t}=S_i - \text{mean}(\{S_i\}_i^G)$~\cite{liu2025drgrpo} where $S_i$ is the reward given to rollout $i$, and $G$ being the number of rollouts for a given prompt and we default to $\alpha=0.5$ and $\beta=5$~\cite{icepop2025}.

We found double-sided masking critical to combat the trainer-inference mismatch: Even when $\pi_{\textcolor{red}{\text{infer}}}$ and $\pi_{\textcolor{blue}{\text{train}}}$ and share the same parameters $\theta$, they can produce significantly different token probabilities, leading to unexpected distribution shifts that can cause runs to crash multiple days into the experiments, if not explicitly addressed. This is similar to CISPO~\cite{minimax2025minimaxm1scalingtesttimecompute} (later further validated in~\cite{khatri2025artscalingreinforcementlearning}, however, we use masking instead of clipping to avoid noisy updates that come with excessive importance ratios. We also apply masking to any rollouts if any of its tokens importance ratio falls under a certain threshold (we use 1e-5 for our training).

\begin{figure}[ht]
\centering
\includegraphics[width=1.0\textwidth]{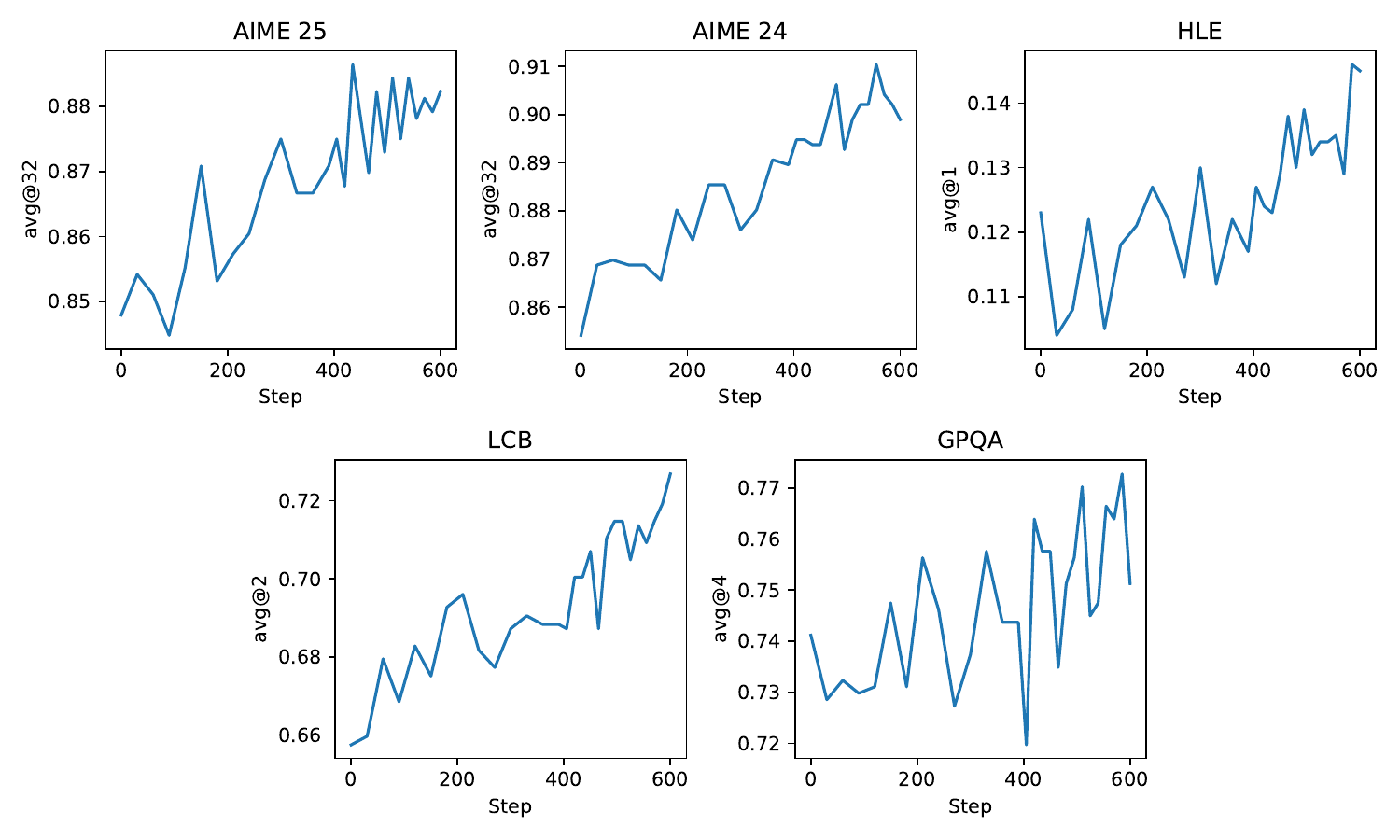}
\captionsetup{width=1.0\textwidth}
\caption{\textbf{Reinforcement Learning.} Reasoning benchmark scores as training progresses. The benchmarks scores generally trend up and do not appear to have reached a plateau.}
\label{training_evals}
\end{figure}

\textbf{Online Evaluation} In Figure \ref{training_evals} we plot the model performance at 15 step intervals on AIME25, AIME24, LiveCodeBench, HLE and GPQA.
The reasoning benchmark scores generally trend up and do not seem to have reached a plateau.
This leads us to believe that allowing the model to continue training would yield continued improvements in the benchmark scores.

\begin{figure}[ht]
\centering
\includegraphics[width=1.0\textwidth]{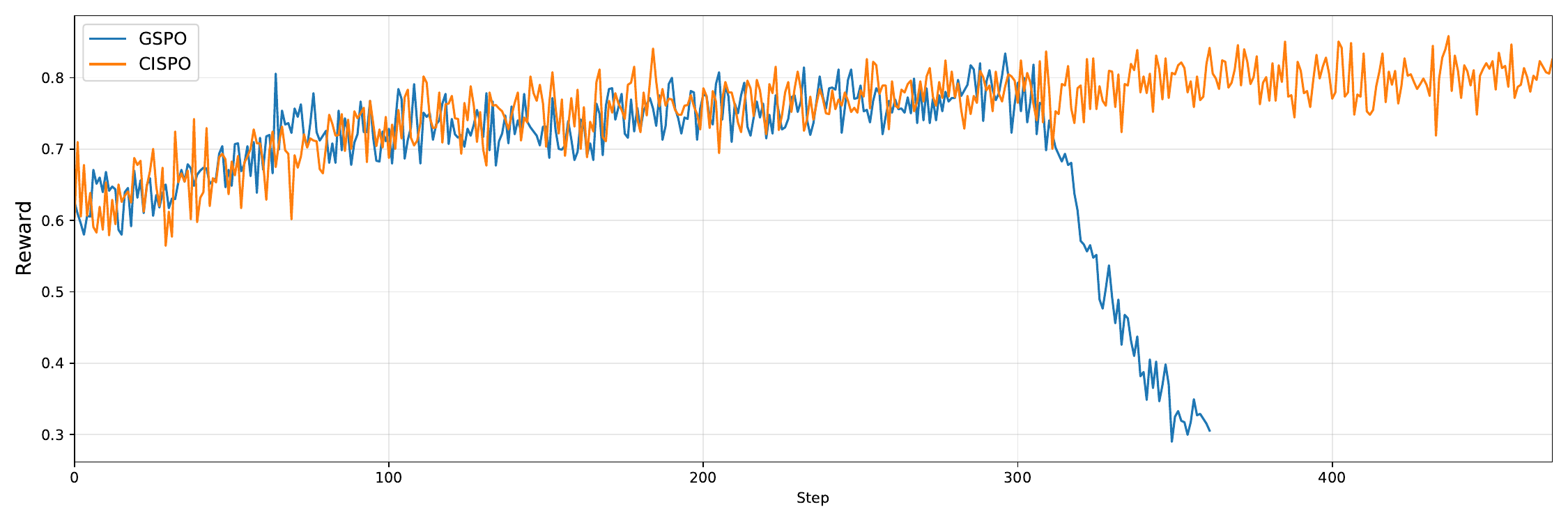}
\captionsetup{width=1.0\textwidth}
\caption{\textbf{Training Stability.} Early ablations of GSPO against CISPO (our algorithm at the time). We use async-8 as a testbed for algorithms to test if they would handle high levels of off-policyness. We observe strange reward (and all other metrics) collapse with GSPO also reported in~\cite{khatri2025artscalingreinforcementlearning}~\cite{qi2025defeatingtraininginferencemismatchfp16}.}
\label{cispo_plot}
\end{figure}

\section{Evaluations}
\label{sec:evaluations}

We evaluate \INTELLECT\ on a wide range of reasoning benchmarks, including AIME 2024\footnote{\pilogo ~\href{https://app.primeintellect.ai/dashboard/environments/primeintellect/aime2024}{primeintellect/aime2024}}, AIME 2025\footnote{\pilogo ~\href{https://app.primeintellect.ai/dashboard/environments/primeintellect/aime2025}{primeintellect/aime2025}}, LiveCodeBench v6\footnote{\pilogo ~\href{https://app.primeintellect.ai/dashboard/environments/primeintellect/livecodebench}{primeintellect/livecodebench}}, GPQA Diamond\footnote{\pilogo ~\href{https://app.primeintellect.ai/dashboard/environments/primeintellect/gpqa}{primeintellect/gpqa}}, HLE\footnote{\pilogo ~\href{https://app.primeintellect.ai/dashboard/environments/primeintellect/hle}{primeintellect/hle}}, MMLU-Pro\footnote{\pilogo ~\href{https://app.primeintellect.ai/dashboard/environments/primeintellect/mmlu-pro}{primeintellect/mmlu-pro}}, 

To ensure a fair comparison, we run evaluations in the exact same settings against API services. Precise details on the evaluation setup are given in Appendix~\ref{sec:reproducing-evaluations}.

\autoref{evals-table} summarizes the results across all benchmarks. \INTELLECT\ outperforms the best matching comparison model, GLM-4.5 Air, which is Z.ai’s post-trained version of the GLM-4.5 Air base model, across every tested benchmark.
Even the $3\times$ larger GLM-4.5 is outperformed on many benchmarks by \INTELLECT, including AIME 2024, AIME 2025 and LiveCodeBench v6.

At the end of RL training, rewards were still increasing with no sign of plateauing in benchmark performance. We will continue training \INTELLECT\  with a focus on agentic environments, to further improve the model's performance on complex agentic tasks.

\begin{table}[t]
\centering
\small
\caption{\centering \textbf{Evaluations.} We report benchmark scores on a wide range of reasoning benchmarks, and compare against models of similar or larger size. All implementations are open-source and reproducible via the \envhub.}
\vspace{5pt}
\begin{tabular}{lccccccc}
\toprule
Benchmark & AIME24 & AIME25 & LCB v6 & GPQA & HLE & MMLU-Pro \\
\midrule
\INTELLECT        & 90.8 & 88.0 & 69.3 & 74.4 & 14.6 & 81.9 \\
GLM-4.5-Air       & 84.6 & 82.0 & 61.5 & 73.3 & 13.3 & 73.9 \\
GLM-4.5           & 85.8 & 83.3 & 64.5 & 77.0 & 14.8 & 83.5\textsuperscript{*} \\
GLM-4.6           & 92.0 & 90.3 & 73.0 & 78.8 &  13.3\textsuperscript{*}    & 83.1  \\
DeepSeek R1 0528  & 83.2 & 73.4 & 62.5 & 77.5 & 15.9 & 75.3 \\
DeepSeek v3.2      & 88.1 & 84.7 & 71.6 & 81.4 & 17.9 & 84.6 \\
GPT-OSS 120B      & 75.8 & 77.7 & 69.9 & 77.3 & 10.6 & 67.1 \\
\bottomrule
\end{tabular}
\label{evals-table}
\end{table}

{
\renewcommand{\thefootnote}{} % Temporarily disable footnote numbering
\footnotetext{\textsuperscript{*} Reported by AA Index}
}

\section{Conclusion \& Future Work}
\label{sec:summary}

In this report, we present \INTELLECT, a 100B+ parameters post-train on top of the GLM-4.5-Air base model. \INTELLECT\ achieves strong performance on math, code, science and broader reasoning benchmarks, and is competitive with or ahead of significantly larger frontier models. The model is trained on a diverse mixture of open environments from the \envhub, including math, code, science, logic, deep research and software engineering tasks, which together target long context tool using agentic behavior.

To support this training run, we introduced a frontier infrastructure stack for reinforcement learning at scale. \primerl\ provides a production grade asynchronous RL framework with disaggregated trainer and inference, continuous batching, in flight weight updates and efficient support for Mixture-of-Experts models. The \verifiers\ library and the \envhub\ standardize how environments and evaluations are expressed, turning them into reusable, versioned artifacts that can be shared, mixed and reproduced across projects. \textit{Prime Sandboxes} and make it possible to execute untrusted code at very high throughput on thousands of concurrent rollouts, and to sustain long multi-week training runs on a 512 H200 cluster.

By open sourcing \INTELLECT, the environments, and the complete training framework, we aim to narrow the gap between proprietary RL pipelines and what independent researchers, small labs and companies can build. The same code that produced \INTELLECT\ is available for single node experiments, mid scale research runs, and for production-scale training. Our hope is that this stack becomes a common foundation for the next generation of open reasoning and agentic models.

There are several directions for future work.

\begin{itemize}
\item \textbf{Scaling Agentic RL.}
By the end of our current RL run, reward and evaluation curves had still not flattened, and the training remained extremely stable. This suggests that \INTELLECT\ is very much still in the high return regime of additional RL compute. With more agentic environments such as DeepDive and Software Engineering in the mix, we expect substantial further gains from simply continuing to train inside these RL environments, particularly for complex agentic use cases.
\item \textbf{Richer RL environments.}  
Over the last few months, more than 500 RL environments have been released on the Environments Hub, and we’ve continued scaling this through our RL Residency and Bounty programs. These environments cover autonomous AI research, computer use, theorem proving, browser automation, and many domain-specific tasks such as law, finance, and tax. \INTELLECT\ used only a small slice of what’s already available. A major next step is scaling RL across a much broader set of high-quality community-contributed environments, covering more tools, modalities, and real-world workloads.
\item \textbf{Long Horizon Agents.} A key next step on our research roadmap is making long horizon behavior RLable by letting the model manage its own context. We've been exploring simple tools for cutting context, prompting itself in isolated sub branches, and maintaining an external memory across turns. These keep the scaffold minimal and let the model learn end-to-end context handling through RL, with future work scaling training on environments that reward effective long-horizon reasoning. In line with recent evidence of “context rot” in long-context models, where the effective reasoning window is much smaller than the advertised context size and performance degrades on long-range reasoning tasks despite successful retrieval of relevant spans~\cite{hiew2025longcontext}, we treat the context window as a scarce resource to be actively managed rather than a passive, ever growing transcript.

\end{itemize}

\newpage
\bibliography{references}
\bibliographystyle{plain}

%%%%%%%%%%%%%%%%%%%%%%%%%%%%%%%%%%%%%%%%%%%%%%%%%%%%%%%%%%%%

\newpage
\appendix

\section{Reproducing Evaluations}
\label{sec:reproducing-evaluations}

\subsection{Evaluation Environments}
\label{subsec:evaluation-environments}

\textbf{MATH-500.} MATH-500 consists of 500 high-school competition math problems distilled from the original MATH~\cite{hendrycks2021math} dataset. To verify responses, we parse reasoning content, extract the final answer from the last \texttt{\textbackslash boxed\{...\}}, and use \texttt{math-verify} to compare the answer against the ground truth. We run two generations per problem, for a total of 1000 generations.

\textbf{AIME.} We evaluate both AIME 2024 and AIME 2025, which each consist of 30 challenging high-school competition math problems. Similar to the MATH 500 environment, we parse reasoning content, extract the final answer from the last \texttt{\textbackslash boxed\{...\}}, and use \texttt{math-verify} to verify responses. We do not employ a LLM-judge for verification thus our reported numbers are more conservative compared to e.g. Artificial Analysis Index, which uses LLM judges. To obtain robust results, we report Avg@32 (Pass@1 over 32 generations per question).

\textbf{GPQA.} GPQA is a Ph.D.-level STEM MCQA benchmark. We use the diamond subset, which includes the 198 hardest questions. We ask the model to put the letter of the final answer in a box, and judge the response by looking for an exact match with the ground truth answer. To obtain robust results we report Avg@4 (Pass@1 over 4 generations per question). 

\textbf{LiveCodeBench.} LiveCodeBench (LCB) is a single-turn coding evaluation benchmark that collects new problems over time from popular programming contests. We use version v6 and include the 454 latest problems (August 2024 to May 2025) as reported in the official \href{https://livecodebench.github.io/leaderboard.html}{LiveCodeBench leaderboard} at the time of writing. We copy the the verification logic from the official \href{https://github.com/LiveCodeBench/LiveCodeBench}{GitHub repository} but integrate it with our sandboxes to ensure secure and scalable test verification. We report Avg@2 (Pass@1 with 2 rollouts per problem) 

\textbf{MMLU-Pro.} MMLU-Pro~\cite{wang2024mmlupro} is a challenging subset of 12K general STEM MCQA questions from MMLU. We ask the model to put the letter of the final answer in a box, and grade by checking for an exact match with the ground truth answer. We report Avg@1 over the 12K samples.

\textbf{HLE.} Humanity's Last Exam (HLE)~\cite{phan2025humanitysexam} consists of 2,500 questions across dozens of subjects, including mathematics, humanities, and the natural sciences. HLE is developed globally by subject-matter experts and consists of multiple-choice and short-answer questions suitable for automated grading. We use the text-only subset, which includes 2,158 examples, problems and do not give the models additional tools. We report the average solve rate across all samples.

\subsection{API models}

\textbf{GLM 4.5 Air, GLM 4.5 \& GLM 4.6} We evaluate \texttt{GLM-4.5-Air}, \texttt{GLM-4.5}~\cite{5team2025glm45agenticreasoningcoding} and \texttt{GLM 4.6} via OpenRouter and enforce that requests are routed to the official z-AI API. We adopt the sampling parameters setup recommended by z-AI. (e.g. we use temperature 0.6 across all benchmarks.)

\textbf{DeepSeek.} We evaluate \texttt{DeepSeek R1 0528} via OpenRouter and enforce that requests are routed to the official DeepSeek API. We did not manage to evaluate \texttt{DeepSeek v3.2 (Thinking)} via OpenRouter as most providers (including the official DeepSeek provider on OpenRouter) would host the chat version. For this reason, we fall back to using the official DeepSeek API with the \texttt{deepseek-reasoner} model slug.

\textbf{OpenAI.} We tried evaluating \texttt{GPT-OSS 120B (High)}~\cite{openai2025gptoss120bgptoss20bmodel} via OpenRouter but from the average response length it was clear that we were evaluating the model with lower than advertised reasoning efforts. We resorted to evaluating \texttt{GPT-OSS 120B (High)} via TogetherAI, where we confirmed that the hosted version is indeed using high reasoning effort. Our observed scores differ slightly from those in the OpenAI model card, and reflect an apples-to-apples comparison from running all models in the same evaluation harnesses, where formatting and scoring logic may differ slightly from other implementations.

\end{document}